\documentclass[10pt,conference]{IEEEtran}

\newif\iffinal
\finaltrue

\newif\ifieee
\ieeefalse

\newif\ifarxiv
\arxivtrue

\ifieee
    \ifarxiv
        \PackageError{Configuration}
        {IEEE and arXiv cannot both be enabled}
        {Set either \string\ieeefalse\space or \string\arxivfalse.}
    \fi
\fi

\usepackage{cite}

\usepackage{graphicx}
\graphicspath{{figs/}}
\DeclareGraphicsExtensions{.pdf,.png,.jpeg}

\usepackage[cmex10]{amsmath}
\usepackage{amsthm}

\usepackage{algorithmic}
\usepackage{array}

\ifCLASSOPTIONcompsoc
  \usepackage[caption=false,font=normalsize,labelfont=sf,textfont=sf]{subfig}
\else
  \usepackage[caption=false,font=footnotesize]{subfig}
\fi

\newcommand{\cmtid}{232}

\iffinal
\else
\usepackage[switch]{lineno}
\fi

\newcommand*{\RL}[2][]{\textcolor{Rhodamine}{[\textbf{\ifthenelse{\equal{#1}{}}{RL}{RL(#1)}}: #2]}}

\usepackage[hyphens]{url}

\usepackage[colorlinks=true,allcolors=black]{hyperref} %

\usepackage[capitalise]{cleveref} %

\usepackage[acronym]{glossaries}
\glsdisablehyper
\usepackage{xspace}
\usepackage{booktabs}
\usepackage[dvipsnames]{xcolor}%
\usepackage[utf8]{inputenc}
\usepackage[T1]{fontenc}
\usepackage[capitalise]{cleveref} %
\usepackage{multirow}

\newcommand\major[1]{#1} %

\newacronymstyle{long-short-br}
{%
  \GlsUseAcrEntryDispStyle{long-short}%
}%
{%
  \GlsUseAcrStyleDefs{long-short}%
}
\setacronymstyle{long-short-br}

\usepackage{transparent}
\usepackage{tikz}
\ifarxiv
    \newcommand\copyrighttext{%
      \scriptsize Accepted for presentation at SIBGRAPI 2026. The final published version will be available on IEEE~Xplore.}
    \newcommand\copyrightnotice{%
    \begin{tikzpicture}[remember picture,overlay]
    \node[anchor=south,yshift=30pt,xshift=0pt] at (current page.south) {\fbox{\transparent{0.85}\parbox{\dimexpr0.6\textwidth-\fboxsep-\fboxrule\relax}{\copyrighttext}}};
    \end{tikzpicture}%
    }
\else
\fi

\ifieee
\IEEEoverridecommandlockouts
\IEEEpubid{\makebox[\columnwidth]{979-8-3195-0255-1/26/\$31.00~\copyright2026 IEEE \hfill}
\hspace{\columnsep}\makebox[\columnwidth]{ }}
\else
\fi

\begin{document}

\title{Benchmarking Spatial, Spectral, and Self-Supervised Cues for Face Forgery Detection under Realistic Degradation}

\iffinal
    \author{
    \IEEEauthorblockN{Lucas Cunha\IEEEauthorrefmark{1}, Lucas Sotomaior\IEEEauthorrefmark{1}, Lucas Gasperin\IEEEauthorrefmark{1},\\Beatriz Caldas\IEEEauthorrefmark{1}, Eduardo Pianovski\IEEEauthorrefmark{1}, and Rayson~Laroca\IEEEauthorrefmark{1}\\[0.75ex]}
    \IEEEauthorblockA{
        \IEEEauthorrefmark{1}\hspace{0.15mm}Pontifical Catholic University of Paran\'a, Curitiba, Brazil\\[0.75ex]
            \hspace{-0.75mm}\IEEEauthorrefmark{1}\hspace{-0.35mm}\tt{\small{\{c.oliveira25,lucas.sotomaior,lucas.gasperin,beatriz.caldas,eduardo.pianovski\}}@pucpr.edu.br} \\ \IEEEauthorrefmark{1}{\tt\small rayson@ppgia.pucpr.br}}
    }
\else
  \author{SIBGRAPI Paper ID: \cmtid \\[7ex]}
  \linenumbers
\fi

\maketitle

\ifarxiv
    \copyrightnotice
\else
\fi

\iffinal
    \newcommand{\urlSupplementary}{\url{https://github.com/lucasdocunha/FaceForgery-Benchmark/}}
\else
    \newcommand{\urlSupplementary}{\textit{[hidden for review]}}
\fi

\newacronym{acc}{ACC}{accuracy}
\newacronym{aigc}{AIGC}{Artificial Intelligence-Generated Content}
\newacronym{auc}{ROC-AUC}{Area Under the ROC Curve}
\newacronym{cnn}{CNN}{Convolutional Neural Network}
\newacronym{fft}{FFT}{Fast Fourier Transform}
\newacronym{gan}{GAN}{Generative Adversarial Network}
\newacronym{gradcam}{Grad-CAM}{Gradient-weighted Class Activation Mapping}
\newacronym{mffi}{MFFI}{Multi-Dimensional Face Forgery Image}
\newacronym{rgb}{RGB}{red-green-blue}
\newacronym{vit}{ViT}{Vision Transformer}

\begin{abstract}
Face forgery detectors often achieve strong results on controlled benchmarks, but their reliability under realistic image degradations remains limited. This paper presents a standardized benchmark for face forgery detection using the \gls*{mffi} dataset and evaluates performance on both clean and degraded test partitions. We compare six model families, including convolutional networks, transformer-based models, and a frozen self-supervised DINOv3 backbone, across spatial, spectral, and hybrid input representations. The results show that clean-set performance is not a reliable indicator of robustness under compression, resizing, and blurring. \major{Xception with RGB obtains the best clean performance, reaching 0.884 mean ROC-AUC}, but degrades substantially on the harder partition. In contrast, frozen DINOv3 achieves the strongest degraded-set result, with \major{0.726 mean ROC-AUC}, while training only a linear classification head. The representation analysis indicates that Fourier-domain cues are most useful when combined with RGB information, whereas purely spectral inputs consistently underperform spatial representations. Qualitative attribution maps further suggest that convolutional detectors focus on localized artifacts, while DINOv3 relies on broader facial structure. These findings reinforce the need for degraded evaluation protocols and highlight self-supervised visual representations as a promising direction for robust face forgery detection.
Our source code is publicly available at \textit{\urlSupplementary}.
\end{abstract}

\IEEEpeerreviewmaketitle

\section{Introduction}

\glsresetall
\glsunset{gradcam}
\glsunset{rgb}

The spread of \gls*{aigc} has made synthetic faces and identity manipulation increasingly realistic and accessible~\cite{lyu_theconversation_2025,milmo_2024,theconversation_2026}.
\glspl*{gan}~\cite{goodfellow2020generative} and diffusion-based models, which are also used for high-fidelity facial reconstruction~\cite{santos2026robust}, can generate photorealistic facial content that threatens privacy, biometric security, and information integrity~\cite{ivanovska2024vulnerability}.
Automated face forgery detection is therefore essential, yet its practical reliability remains limited, motivating evaluations that go beyond classification performance~\cite{anlen_vazquezllorente_2024,lopes2025alem}.

Most detectors are trained as binary classifiers that learn spatial artifacts, temporal inconsistencies, or frequency irregularities.
On common benchmarks, \glspl*{cnn} and \glspl*{vit} can reach \gls*{auc} values above $0.99$~\cite{ref_generalization_gap}.
These results, however, often rely on controlled data distributions and can degrade sharply when test images differ from training images~\cite{ref_sota_detection}.
This gap is critical because images shared on online platforms are routinely compressed, resized, blurred, or re-encoded, which may suppress the subtle traces that forensic models rely~on.

The \gls*{mffi} dataset~\cite{ref_mffi} was recently introduced to study this gap through more than 50 forgery techniques, diverse authentic sources, and multi-level degradations designed to approximate real-world conditions.
It enables evaluation on both a clean test partition and a degraded partition, making it well-suited for measuring whether performance under laboratory conditions transfers to realistic conditions.
Rather than proposing a single new detector, this paper uses \gls*{mffi} to establish a standardized benchmark across architectures and input~representations.

We evaluate six model families that cover lightweight and residual convolutional models, attention-based models, and self-supervised foundation features.
The benchmark includes MobileNetV3~\cite{howard2019searching}, Xception~\cite{chollet2017xception}, ResNet-18~\cite{ref_resnet}, a \gls*{vit}~\cite{ref_vit}, a CLIP-style encoder~\cite{ref_clip}, and a frozen DINOv3 backbone~\cite{simeoni2025dinov3}.
For compatible architectures, we additionally compare RGB, Fourier-based spectral representations, and hybrid spatial-spectral stacks.
This design separates the contributions of the model family and the input representation while keeping the evaluation protocol fixed.

The contributions of this paper are threefold:
\begin{itemize}
    \item We provide a standardized benchmark of six heterogeneous architectures on MFFI, reporting performance on both clean and degraded partitions under a common protocol. Our code is publicly available at \textit{\urlSupplementary}, supporting reproducibility beyond the reported~results;
    \item We quantify the effect of seven spatial, spectral, and hybrid input representations, showing when frequency cues help and when they fail;
    \item We compare our results with state-of-the-art results on \gls*{mffi} and use attribution methods~\cite{selvaraju2017gradcam,abnar2020quantifying} to analyze the evidence used by accurate and degraded~detectors.
\end{itemize}

\section{Related Work}
\label{sec:related}

\subsection{Detection of Synthetic and Manipulated Faces}

Early face forgery detectors were commonly trained end-to-end on curated benchmarks and reached near-perfect in-distribution accuracy.
R\"ossler et al.~\cite{ref_faceforensics} introduced FaceForensics++ and showed that Xception~\cite{chollet2017xception} could identify several facial manipulations effectively.
Wang et al.~\cite{ref_generalization_gap} showed that a single \gls*{cnn} trained on one generator can generalize to unseen generators when paired with appropriate augmentation, but also showed that performance decreases when the test distribution changes.
Recent surveys reinforce this limitation, reporting that detectors often learn generator-specific or post-processing-specific cues that transfer poorly across datasets and acquisition conditions~\cite{ref_sota_detection}.
Ivanovska and \v{S}truc~\cite{ivanovska2024vulnerability} further showed that denoising-diffusion attacks can suppress detector evidence without perceptually obvious changes.
Our work follows this robustness-oriented view and evaluates how several detector families behave under the explicit degradation protocol available in \gls*{mffi}.

\subsection{Frequency-Domain Forensics}

\begin{figure*}[!htb]
\centering
\includegraphics[width=0.8\linewidth]{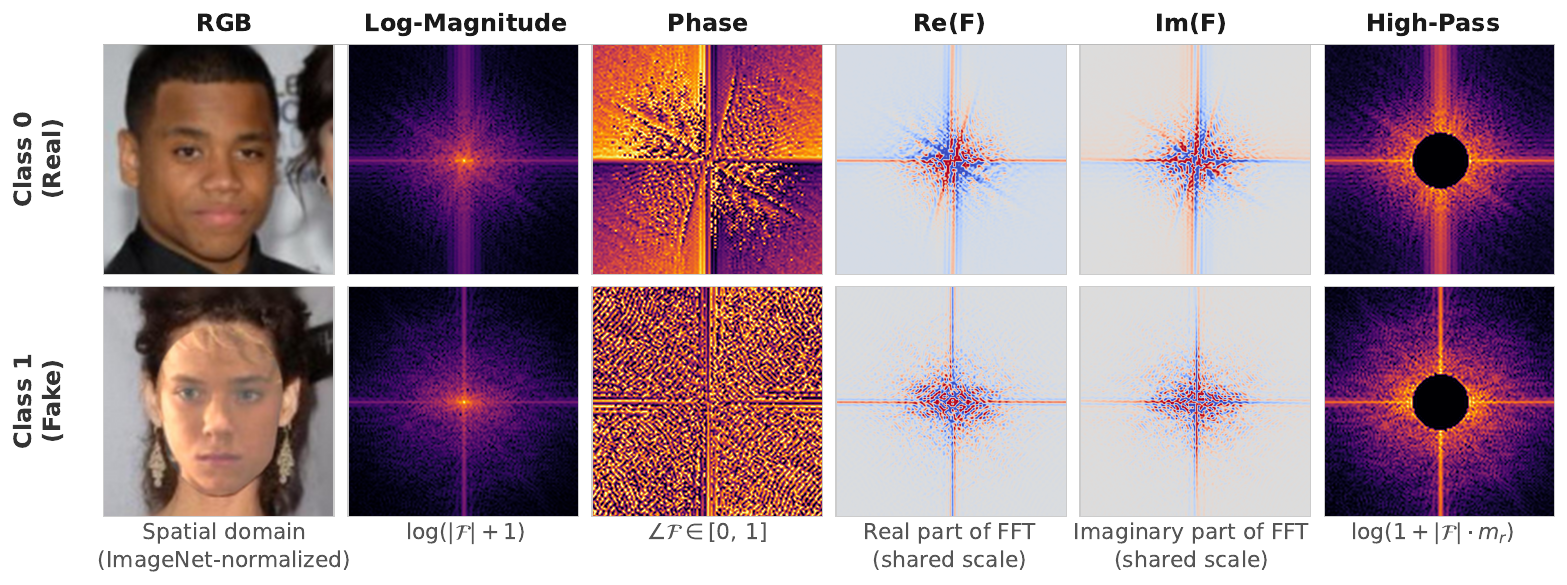}

\vspace{-2mm}

\caption{Spatial and frequency-domain views of authentic and forged face images.
Each row shows the RGB input, log-magnitude spectrum, phase, real and imaginary \gls*{fft} components, and high-pass magnitude.}
\label{fig:fourier_transformacoes}
\end{figure*}

Frequency-domain forensics is motivated by the observation that generative pipelines may leave structured spectral traces.
Frank et al.~\cite{frank2020leveraging} reported systematic Fourier artifacts in \gls*{gan}-generated images, especially at high frequencies.
Dzanic et al.~\cite{dzanic2020fourier} observed measurable spectrum discrepancies between deep network-generated images and natural photographs.
Qian et al.~\cite{qian2020thinking} proposed a frequency-aware detector that mines clues from multiple frequency bands.
These studies motivate our representation axis: instead of testing a single spectral input, we compare seven spatial, spectral, and hybrid encodings under the same training and evaluation protocol.
\cref{fig:fourier_transformacoes} illustrates these encodings on authentic and forged faces.

\subsection{Foundation Models and Self-Supervised Learning}

Large pre-trained encoders offer an alternative to learning forensic evidence solely from the target dataset.
CLIP~\cite{ref_clip} learns transferable visual representations through image-text supervision, while DINOv2~\cite{ref_dino} and DINOv3~\cite{simeoni2025dinov3} learn strong visual features through self-supervision.
Recent robust deepfake detection systems follow this direction: the winning~\cite{qu2026dino_mac} and runner-up~\cite{wu2026loger} solutions of the CVPR 2026 Robust DeepFake Detection Challenge both use adapted DINOv3 backbones.
We include frozen DINOv3 features to assess how much robustness is available without backbone fine-tuning, ensembling, or additional task-specific~data.

\section{Experimental Setup}
\label{sec:experimental_setup}

This section details the dataset partitions, input representations, model families, training procedure, and metrics used in the~benchmark.

\subsection{Dataset and Partitions}

\begin{figure}[!htb]
\centering
\includegraphics[width=0.22\columnwidth]{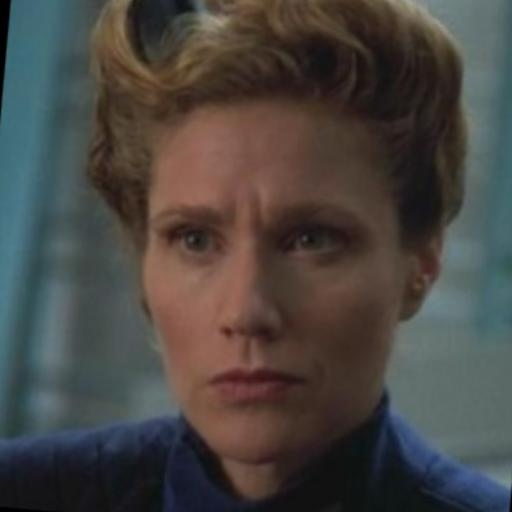}
\includegraphics[width=0.22\columnwidth]{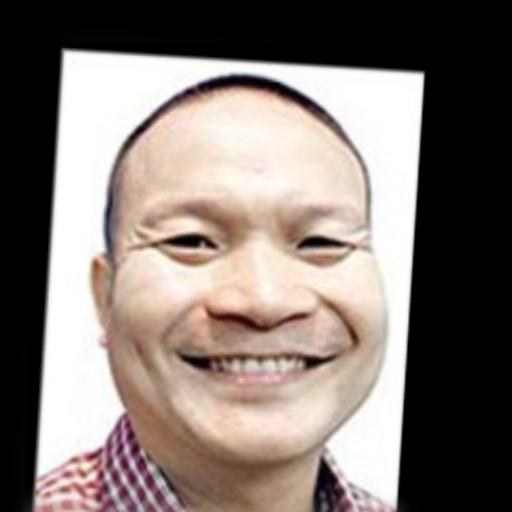}
\includegraphics[width=0.22\columnwidth]{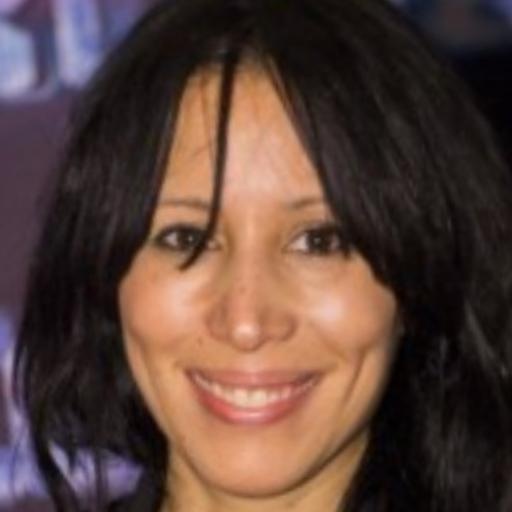}
\includegraphics[width=0.22\columnwidth]{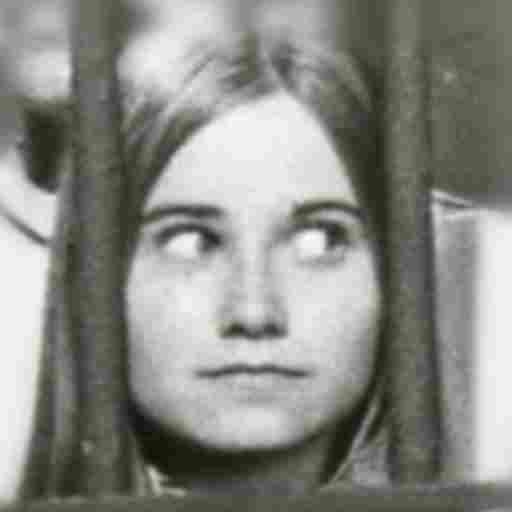}

\vspace{1mm}

\includegraphics[width=0.22\columnwidth]{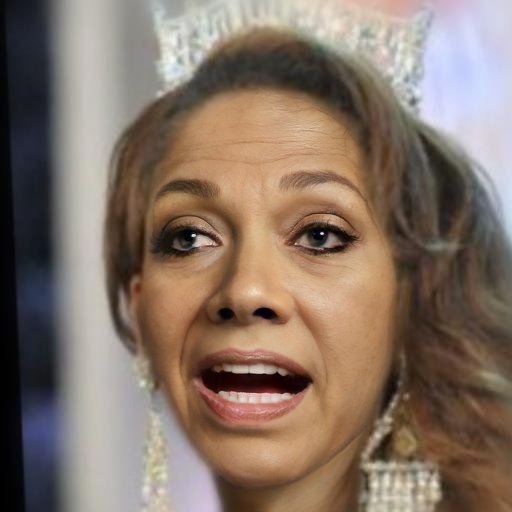}
\includegraphics[width=0.22\columnwidth]{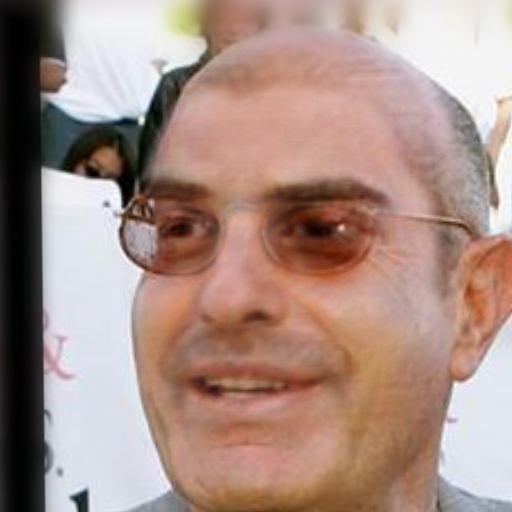}
\includegraphics[width=0.22\columnwidth]{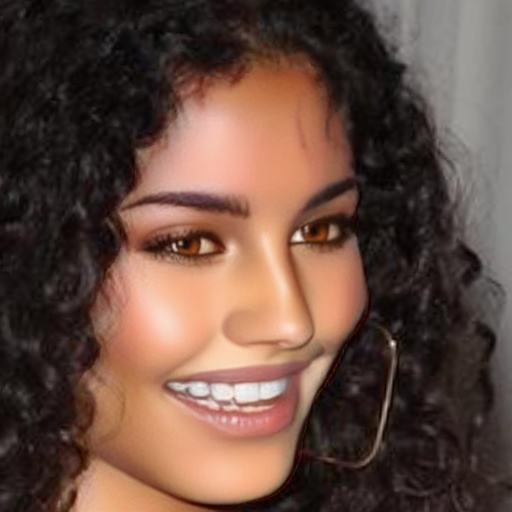}
\includegraphics[width=0.22\columnwidth]{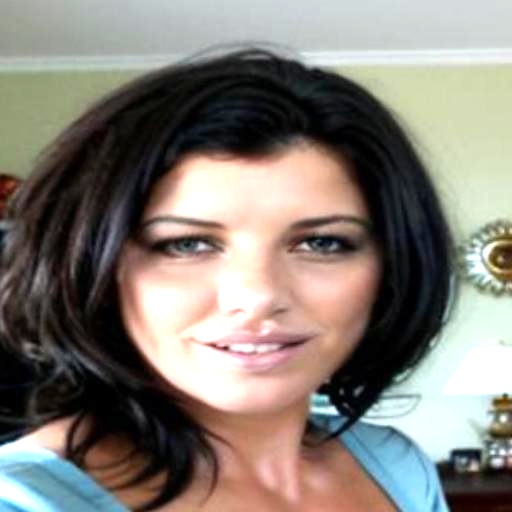}

\vspace{-1mm}

\caption{Examples of authentic (top) and forged (bottom) face images from the \gls*{mffi} benchmark~\cite{ref_mffi}.}
\label{fig:real_fake}
\end{figure}

All experiments use the \gls*{mffi} dataset~\cite{ref_mffi}, which combines diverse forgery methods, facial scenes, authentic sources, and degradation operations.
The task is binary image classification, with each sample labeled as authentic or forged, as illustrated in \cref{fig:real_fake}.
We use the official splits with $524{,}429$ training images, $147{,}363$ validation images, and $181{,}947$ test images.
The test partition is moderately imbalanced, with approximately $57\%$ forged samples.
Robustness is evaluated on \emph{Test-Hard}, the degraded version of the same test set, which corresponds to Test-D in the original \gls*{mffi} benchmark~\cite{ref_mffi}.
This paired evaluation exposes the difference between clean benchmark performance and performance after compression, resizing, and~blurring.

\subsection{Input Representations}
\label{sec:fourier_modes}

Each image is decoded as RGB.
For spectral modes, the image is converted to grayscale luminance, $Y = 0.299R + 0.587G + 0.114B$, before a centered two-dimensional \gls*{fft}.
From the spectrum, we derive log-magnitude, phase, real and imaginary components, and a high-pass magnitude that suppresses the low-frequency disk.
\cref{tab:modes} summarizes the seven evaluated representations.
When the number of channels differs from three, the first convolutional or patch-embedding layer is adapted by reinitializing the corresponding channel weights.
Spatial augmentations are applied before computing the Fourier transforms, preserving alignment between image content and spectral channels.

\begin{table}[!htb]
\centering
\caption{Input representations evaluated in the benchmark.}
\label{tab:modes}
\vspace{-2mm}
\resizebox{0.98\linewidth}{!}{
\begin{tabular}{@{}llc@{}}
\toprule
Mode & Description & Ch. \\
\midrule
\textit{RGB} & spatial image & 3 \\
\textit{Log-Magnitude} & $\log(|\mathcal{F}|+1)$ & 1 \\
\textit{Phase} & phase mapped to $[0,1]$ & 1 \\
\textit{Complex Spectrum} & real and imaginary spectra & 2 \\
\textit{RGB+Mag} & \gls*{rgb} plus log-magnitude & 4 \\
\textit{High-Pass Spectrum} & high-pass magnitude & 1 \\
\textit{RGB+Freq} & \gls*{rgb} plus magnitude, phase, and high-pass & 6 \\
\bottomrule
\end{tabular}
}
\end{table}

The \textit{Log-Magnitude} mode is normalized to $[0,1]$ and captures the energy distribution across spatial frequencies.
The \textit{Phase} mode maps angles from $[-\pi,\pi]$ to $[0,1]$, preserving structural information.
The \textit{Complex Spectrum} mode keeps real and imaginary components, each normalized by the maximum absolute value.
The \textit{High-Pass Spectrum} mode removes a circular low-frequency region with radius $r = 0.12\min(H,W)$, emphasizing the high-frequency artifacts discussed in prior work~\cite{frank2020leveraging}.
The hybrid modes concatenate spatial and spectral channels.

\subsection{Architectures}

We evaluate six model families spanning convolutional, attention-based, and self-supervised paradigms.
\cref{tab:arch} lists the input size and pre-training setting for each family.
Xception, ResNet-18, MobileNetV3, the \gls*{vit}, and the CLIP-style encoder are trained from scratch, so their results reflect what each architecture learns from \gls*{mffi} alone.
DINOv3 is the only pre-trained model, and its backbone remains frozen while a linear head is trained.

\begin{table}[!htb]
\centering
\caption{Architectures evaluated in the benchmark.}
\label{tab:arch}
\vspace{-2mm}
\resizebox{0.98\linewidth}{!}{
\begin{tabular}{@{}llcc@{}}
\toprule
Model & Family & Input & Pre-training \\
\midrule
Xception~\cite{chollet2017xception} & separable \gls*{cnn} & $224\times224$ & none \\
ResNet-18~\cite{ref_resnet} & residual \gls*{cnn} & $224\times224$ & none \\
MobileNetV3~\cite{howard2019searching} & lightweight \gls*{cnn} & $224\times224$ & none \\
ViT~\cite{ref_vit} & transformer & $224\times224$ & none \\
CLIP enc.~\cite{ref_clip} & transformer & $224\times224$ & none \\
DINOv3~\cite{simeoni2025dinov3} & self-supervised transformer & $224\times224$ & frozen \\
\bottomrule
\end{tabular}
}
\end{table}

The \gls*{vit} uses $16\times16$ patches, hidden dimension
$128$, three encoder layers, four attention heads, Mixup with
$\alpha=0.2$, and dropout $0.25$.
The CLIP-style encoder follows the image-side design with hidden dimension $256$, six encoder
layers, eight attention heads, and projection dimension $128$,
but does not load CLIP weights.
The DINOv3 classifier consists of LayerNorm, Dropout, and Linear layers on top of the frozen~backbone.

\subsection{Training and Evaluation}

All models are optimized with AdamW and a cross-entropy loss.
We use weight decay $10^{-4}$, gradient clipping at norm $1.0$, mixed precision, early stopping, and a ReduceLROnPlateau scheduler.
Class imbalance is handled with a weighted random sampler.
Data augmentation includes random resized crops, horizontal flips, and color jitter before \gls*{fft} computation.
\major{Results are reported as the mean and standard deviation across three
independent runs with different random seeds.}

The decision threshold is selected on the validation set and fixed for Test and Test-Hard.
We report \gls*{auc} as the primary metric because it is threshold-independent and well suited to the class imbalance in \gls*{mffi}.
We also report \gls*{acc} at this threshold.
For qualitative analysis, we apply Grad-CAM~\cite{selvaraju2017gradcam} to the convolutional models and DINOv3, and Attention Rollout~\cite{abnar2020quantifying} to the ViT and CLIP-style~encoder.

\section{Results and Discussion}
\label{sec:results}

\begin{figure}[!htb]
\centering
\includegraphics[width=0.86\columnwidth]{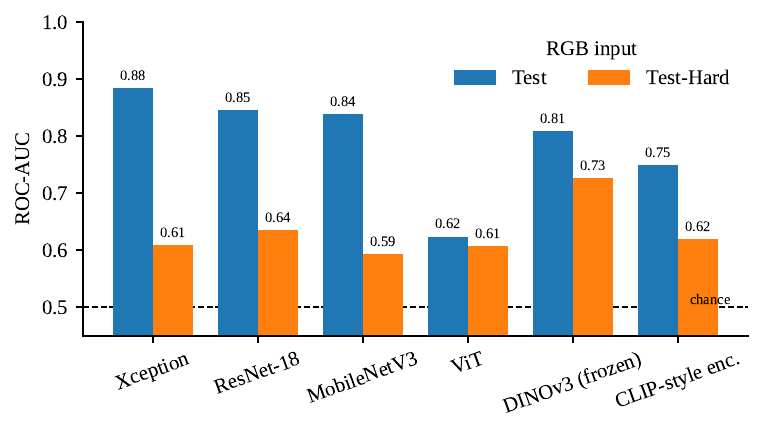}
\vspace{-3mm}
\caption{\gls*{rgb} ROC-AUC on the clean Test and degraded Test-Hard partitions.}
\label{fig:gap}
\end{figure}

\major{\cref{tab:results} reports \gls*{auc} and \gls*{acc} for every architecture and input representation on Test and Test-Hard, while \cref{fig:gap} illustrates the performance gap between Test and Test-Hard across models.}
The benchmark reveals three main patterns: clean performance favors task-specific convolutional models, degradation changes the model ranking, and frequency information is useful mainly as an auxiliary signal.

\begin{table}[!htb]
\centering
\caption{\major{\gls*{auc} and accuracy on the clean Test and degraded Test-Hard partitions of \gls*{mffi}. Best results are shown in~\textbf{bold}.}}
\label{tab:results}
\vspace{-2mm}
\resizebox{0.99\linewidth}{!}{
\begin{tabular}{@{}ll cc cc@{}}
\toprule
 & & \multicolumn{2}{c}{\major{Test (clean)}} & \multicolumn{2}{c}{\major{Test-Hard (degraded)}} \\
\cmidrule(lr){3-4}\cmidrule(lr){5-6}
Model & Representation & \gls*{auc} & ACC & \gls*{auc} & ACC \\
\midrule
Xception~\cite{chollet2017xception} & \textit{RGB} & \textbf{0.884 $\pm$ 0.0091} & 0.757 $\pm$ 0.0051 & 0.609 $\pm$ 0.0196 & 0.568 $\pm$ 0.0033 \\
 & \textit{Log-Magnitude} & 0.686 $\pm$ 0.0125 & 0.638 $\pm$ 0.0090 & 0.517 $\pm$ 0.0200 & 0.535 $\pm$ 0.0292 \\
 & \textit{Phase} & 0.802 $\pm$ 0.0228 & 0.731 $\pm$ 0.0174 & 0.573 $\pm$ 0.0031 & 0.571 $\pm$ 0.0030 \\
 & \textit{Complex Spectrum} & 0.790 $\pm$ 0.0274 & 0.720 $\pm$ 0.0266 & 0.608 $\pm$ 0.0282 & 0.577 $\pm$ 0.0238 \\
 & \textit{RGB+Mag} & 0.878 $\pm$ 0.0152 & 0.768 $\pm$ 0.0115 & 0.650 $\pm$ 0.0231 & 0.603 $\pm$ 0.0189 \\
 & \textit{High-Pass Spectrum} & 0.674 $\pm$ 0.0196 & 0.622 $\pm$ 0.0116 & 0.501 $\pm$ 0.0174 & 0.520 $\pm$ 0.0132 \\
 & \textit{RGB+Freq} & 0.865 $\pm$ 0.0038 & \textbf{0.777 $\pm$ 0.0105} & 0.609 $\pm$ 0.0536 & 0.580 $\pm$ 0.0329 \\
\midrule
ResNet-18~\cite{ref_resnet} & \textit{RGB} & 0.846 $\pm$ 0.0201 & 0.725 $\pm$ 0.0238 & 0.635 $\pm$ 0.0001 & 0.573 $\pm$ 0.0001 \\
 & \textit{Log-Magnitude} & 0.703 $\pm$ 0.0289 & 0.647 $\pm$ 0.0210 & 0.533 $\pm$ 0.0003 & 0.547 $\pm$ 0.0002 \\
 & \textit{Phase} & 0.730 $\pm$ 0.0335 & 0.676 $\pm$ 0.0253 & 0.583 $\pm$ 0.0002 & 0.579 $\pm$ 0.0003 \\
 & \textit{Complex Spectrum} & 0.758 $\pm$ 0.0490 & 0.694 $\pm$ 0.0410 & 0.569 $\pm$ 0.0000 & 0.568 $\pm$ 0.0002 \\
 & \textit{RGB+Mag} & 0.841 $\pm$ 0.0226 & 0.742 $\pm$ 0.0211 & 0.634 $\pm$ 0.0003 & 0.576 $\pm$ 0.0000 \\
 & \textit{High-Pass Spectrum} & 0.685 $\pm$ 0.0135 & 0.633 $\pm$ 0.0150 & 0.519 $\pm$ 0.0000 & 0.538 $\pm$ 0.0000 \\
 & \textit{RGB+Freq} & 0.831 $\pm$ 0.0310 & 0.746 $\pm$ 0.0269 & 0.648 $\pm$ 0.0001 & 0.600 $\pm$ 0.0001 \\
\midrule
MobileNetV3~\cite{howard2019searching} & \textit{RGB} & 0.839 $\pm$ 0.0313 & 0.751 $\pm$ 0.0212 & 0.594 $\pm$ 0.0001 & 0.563 $\pm$ 0.0002 \\
 & \textit{Log-Magnitude} & 0.669 $\pm$ 0.0294 & 0.628 $\pm$ 0.0271 & 0.503 $\pm$ 0.0002 & 0.511 $\pm$ 0.0001 \\
 & \textit{Phase} & 0.689 $\pm$ 0.0252 & 0.646 $\pm$ 0.0215 & 0.561 $\pm$ 0.0002 & 0.572 $\pm$ 0.0000 \\
 & \textit{Complex Spectrum} & 0.733 $\pm$ 0.0443 & 0.671 $\pm$ 0.0361 & 0.543 $\pm$ 0.0002 & 0.560 $\pm$ 0.0000 \\
 & \textit{RGB+Mag} & 0.828 $\pm$ 0.0244 & 0.746 $\pm$ 0.0181 & 0.612 $\pm$ 0.0000 & 0.587 $\pm$ 0.0000 \\
 & \textit{High-Pass Spectrum} & 0.653 $\pm$ 0.0245 & 0.622 $\pm$ 0.0082 & 0.519 $\pm$ 0.0497 & 0.520 $\pm$ 0.0330 \\
 & \textit{RGB+Freq} & 0.798 $\pm$ 0.0231 & 0.726 $\pm$ 0.0195 & 0.548 $\pm$ 0.0499 & 0.539 $\pm$ 0.0330 \\
\midrule
ViT~\cite{ref_vit} & \textit{RGB} & 0.624 $\pm$ 0.0412 & 0.581 $\pm$ 0.0069 & 0.608 $\pm$ 0.0273 & 0.581 $\pm$ 0.0069 \\
 & \textit{Log-Magnitude} & 0.551 $\pm$ 0.0684 & 0.577 $\pm$ 0.0215 & 0.550 $\pm$ 0.0692 & 0.577 $\pm$ 0.0215 \\
 & \textit{Phase} & 0.576 $\pm$ 0.0343 & 0.578 $\pm$ 0.0088 & 0.567 $\pm$ 0.0278 & 0.577 $\pm$ 0.0088 \\
 & \textit{Complex Spectrum} & 0.608 $\pm$ 0.0181 & 0.574 $\pm$ 0.0002 & 0.601 $\pm$ 0.0093 & 0.574 $\pm$ 0.0002 \\
 & \textit{RGB+Mag} & 0.645 $\pm$ 0.0629 & 0.597 $\pm$ 0.0144 & 0.630 $\pm$ 0.0482 & 0.595 $\pm$ 0.0159 \\
 & \textit{High-Pass Spectrum} & 0.488 $\pm$ 0.0224 & 0.572 $\pm$ 0.0013 & 0.485 $\pm$ 0.0206 & 0.572 $\pm$ 0.0013 \\
 & \textit{RGB+Freq} & 0.633 $\pm$ 0.0025 & 0.596 $\pm$ 0.0074 & 0.582 $\pm$ 0.0477 & 0.581 $\pm$ 0.0147 \\
\midrule
DINOv3~\cite{simeoni2025dinov3} & \textit{RGB} & 0.809 $\pm$ 0.0046 & 0.712 $\pm$ 0.0158 & \textbf{0.726 $\pm$ 0.0000} & \textbf{0.663 $\pm$ 0.0022} \\
\midrule
CLIP-style enc.~\cite{ref_clip} & \textit{RGB} & 0.750 $\pm$ 0.0043 & 0.683 $\pm$ 0.0015 & 0.619 $\pm$ 0.0006 & 0.581 $\pm$ 0.0008 \\
\bottomrule
\end{tabular}
}
\end{table}

\subsection{Clean and Degraded Performance}

\major{On the clean Test partition, convolutional detectors achieve the strongest results.
Xception obtains the highest \gls*{auc} ($0.884$ with \textit{RGB}), followed by ResNet-18 ($0.846$ with \textit{RGB}) and MobileNetV3 ($0.839$ with \textit{RGB}).
The \gls*{vit} performs substantially worse, with \gls*{acc} close to the majority-class rate and a maximum \gls*{auc} of $0.645$.
Despite training only a linear head, the frozen DINOv3 backbone remains competitive, reaching an \gls*{auc} of~$0.809$.}

\major{The ranking changes markedly under degradation.
With \textit{RGB}, Xception decreases from $0.884$ to $0.609$, while MobileNetV3 decreases from $0.839$ to $0.594$.
The strongest convolutional result on Test-Hard is obtained by Xception with \textit{RGB+Mag} ($0.650$), followed closely by ResNet-18 with \textit{RGB+Freq} ($0.648$).
In contrast, frozen DINOv3 decreases only from $0.809$ to $0.726$ and achieves the highest Test-Hard performance by a clear margin.
This pattern suggests that representations learned exclusively for the target forgery task may overfit fragile generator- or acquisition-specific cues, whereas self-supervised features retain more useful information after compression, resizing, and blurring.}

The \gls*{vit} shows a small absolute drop, but this should not be interpreted as strong robustness.
Its clean performance is already low, so the apparent stability mainly reflects a floor effect rather than invariance to degradation.

\subsection{Spatial, Spectral, and Hybrid Inputs}

\major{Frequency-domain information provides limited benefit on the clean Test set but becomes more useful under Test-Hard degradation.
For every convolutional architecture, at least one hybrid representation, \textit{RGB+Mag} or \textit{RGB+Freq}, matches or outperforms \gls*{rgb} on Test-Hard.
In contrast, purely spectral representations remain substantially weaker on both partitions.
These results indicate that spectral cues do not replace spatial information, but can provide complementary evidence when degradation weakens RGB-based cues.}

\major{One plausible explanation lies in the inductive biases of convolutional architectures.
Image-generation pipelines can leave systematic frequency-domain irregularities, including periodic patterns associated with upsampling and abnormal high-frequency statistics~\cite{frank2020leveraging,dzanic2020fourier,qian2020thinking}.
As \glspl*{cnn} rely on local shared filters and are sensitive to texture-like patterns~\cite{geirhos2019texture}, early concatenation may allow them to combine these spectral cues effectively with RGB information.
The smaller gains obtained by the ViT suggest that this fusion strategy may be better aligned with convolutional models. DINOv3 and the CLIP-style encoder were evaluated only with RGB and are therefore excluded from this~comparison.}

\subsection{Comparison with Published \gls*{mffi} Results}
\label{sec:literature}

\major{\cref{tab:literature} compares our results with image-level \gls*{mffi} results reported in the literature.
The official \gls*{mffi} benchmark is the most comparable protocol because it trains on \gls*{mffi} and reports both the clean and degraded partitions~\cite{ref_mffi}.
MAP-Mamba is evaluated in a cross-dataset setting, with training on FaceForensics++ and testing on MFFI, so it measures transfer rather than in-distribution learning~\cite{ref_mapmamba}.
DeFakerOne is trained on a 12.5M-sample multi-domain collection that includes \gls*{mffi} itself, reflecting a substantially different supervision scale~\cite{guangjian2026venusdefakerone}.}

\begin{table}[!htb]
\centering
\caption{Image-level ROC-AUC on \gls*{mffi} reported in the literature and in this work.
``Test-Hard'' denotes the official degraded test partition, called Test-D in~\cite{ref_mffi}.
Best results within the comparable \gls*{mffi}-training protocol are in \textbf{bold}, and n/r means not reported.}
\label{tab:literature}
\vspace{-2mm}
\resizebox{0.99\linewidth}{!}{
\begin{tabular}{@{}llcc@{}}
\toprule
Method & Training data & Test & Test-Hard \\
\midrule
Xception~(from~\cite{ref_mffi}) & \gls*{mffi} & 0.852 & 0.708 \\
RFM~(from~\cite{ref_mffi}) & \gls*{mffi} & 0.849 & 0.720 \\
SRM~(from~\cite{ref_mffi}) & \gls*{mffi} & 0.877 & 0.580 \\
SPSL~(from~\cite{ref_mffi}) & \gls*{mffi} & 0.846 & 0.714 \\
MAP-Mamba~\cite{ref_mapmamba} & FF++ cross-dataset & \phantom{$^{\dagger}$}0.668$^{\dagger}$ & n/r \\
DeFakerOne~\cite{guangjian2026venusdefakerone} & 12.5M multi-domain & \phantom{$^{\dagger}$}0.961$^{\dagger}$ & n/r \\
\midrule
Xception, \textit{RGB} (ours) & \gls*{mffi} & \textbf{0.884} & 0.609 \\
ResNet-18, \textit{RGB} (ours) & \gls*{mffi} & 0.846 & 0.635 \\
DINOv3 frozen, \textit{RGB} (ours) & \gls*{mffi} & 0.809 & \textbf{0.726} \\
\bottomrule
\end{tabular}
}
\par\smallskip
\raggedright\footnotesize $^{\dagger}$The original paper does not specify whether the evaluation partition is clean or degraded.
\end{table}

\major{Within the comparable protocol, the best results in our benchmark match or improve the published \gls*{mffi} baselines on both partitions.
On the clean partition, Xception with \textit{RGB} reaches an \gls*{auc} of $0.884$, exceeding the strongest official result (SRM, $0.877$) and the official Xception baseline ($0.852$).
On Test-Hard, frozen DINOv3 reaches $0.726$, marginally exceeding the strongest published degraded-set result (RFM, $0.720$), while training only a linear head on a frozen backbone.
Although this margin is small, the result shows that a frozen self-supervised representation can match the strongest task-specific baseline with minimal training.
More broadly, the comparison confirms that the methods leading on clean data are not necessarily those that remain strongest after~degradation.}

Recent challenge results are consistent with this finding.
DINO-MAC~\cite{qu2026dino_mac} and LOGER~\cite{wu2026loger}, the first- and second-place solutions of the \textit{CVPR 2026 Robust DeepFake Detection Challenge}, both build on DINOv3 backbones and obtain strong results under degradation-heavy evaluation.
Our benchmark shows that this advantage is already visible in the frozen representation itself, before fine-tuning, large ensembles, or additional training data are introduced.

\major{The DeFakerOne result of $0.961$ is the highest value in \cref{tab:literature}, but it is not directly comparable with the \gls*{mffi}-only protocol.
Its training mixture includes \gls*{mffi}, contains approximately $24\times$ as many samples, and incorporates proprietary data, while the evaluated partition is not specified.
The result therefore illustrates the potential benefit of broader supervision, rather than providing a controlled comparison with models trained exclusively on \gls*{mffi}.}

\subsection{Qualitative Evidence from Attribution Maps}
\label{sec:gradcam}

\begin{figure}[!htb]
\centering
\includegraphics[width=0.9\linewidth]{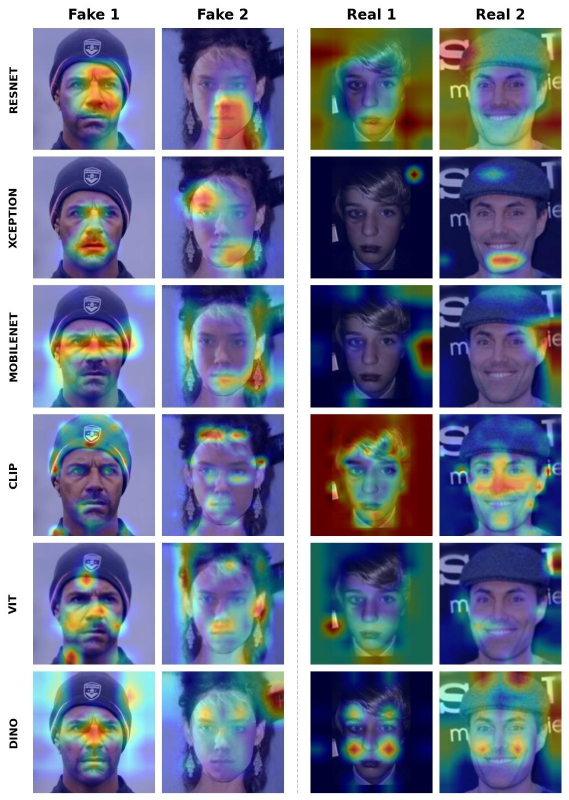}

\vspace{-2mm}

\caption{Attribution maps for the six \gls*{rgb} models on forged (left) and authentic (right) inputs. \gls*{gradcam}~\cite{selvaraju2017gradcam} is applied to ResNet-18, Xception, MobileNetV3, and DINOv3, while Attention Rollout~\cite{abnar2020quantifying} is applied to the ViT and CLIP-style encoder. Convolutional detectors emphasize localized facial regions, from-scratch attention models produce weaker and more diffuse localization, and DINOv3 distributes evidence across broader facial~landmarks.}
\label{fig:gradcam}
\end{figure}

To inspect model evidence, we generate attribution maps for all \gls*{rgb} models on four common samples.
We apply \gls*{gradcam}~\cite{selvaraju2017gradcam} to the convolutional detectors and DINOv3, and Attention Rollout~\cite{abnar2020quantifying} to the from-scratch transformers.
\cref{fig:gradcam} shows two forged and two authentic examples.

These samples are illustrative and do not constitute a quantitative evaluation, but they reveal a consistent qualitative divergence.
All six models classified both forged samples correctly.
For the authentic samples, the convolutional detectors also classified both correctly, whereas the CLIP-style encoder, \gls*{vit}, and DINOv3 classified both as forged, consistent with their broader false-positive behavior.

The attribution maps help explain this divergence.
Convolutional detectors tend to activate on localized facial regions in forged images, such as the nose, mouth, eyes, or blending boundaries.
On authentic images, their maps are more diffuse, suggesting that decisions may rely on the absence of localized~anomalies.

The from-scratch transformers show less consistent evidence.
The \gls*{vit} produces scattered activations that poorly align with common manipulation regions and activates broadly on authentic faces.
The CLIP-style encoder similarly shows limited localization and high activation on authentic samples, consistent with its false-positive errors.

DINOv3 exhibits a different pattern: its evidence is distributed across multiple semantically meaningful facial landmarks rather than concentrated in a single region.
This broader allocation is consistent with its robustness on Test-Hard, but may also reduce discrimination on clean authentic samples, contributing to the observed misclassifications.

\subsection{Limitations}
\label{sec:limitations}

\major{Four limitations constrain the conclusions of this study.
First, all experiments use a single dataset, so the benchmark measures robustness to the \gls*{mffi} degradation protocol rather than cross-dataset generalization; the reported ranking should therefore be validated on additional forgery sources.
Second, five of the six model families are trained from scratch. This isolates what each architecture learns from \gls*{mffi}, but does not estimate the performance attainable with large-scale pretraining.
Third, DINOv3 is evaluated only as a frozen feature extractor with a linear head, so the experiments do not establish whether its robustness advantage persists after fine-tuning.
Finally, the attribution analysis is qualitative and correlational, and is based on only four samples. Moreover, applying \gls*{gradcam} to a frozen transformer backbone requires reshaping tokens into a spatial grid, making the resulting maps approximate visualizations rather than exact gradient attributions.
Accordingly, we use these maps only to support interpretation of \cref{tab:results}, not as independent evidence for the quantitative conclusions.}

\section{Conclusions}
\label{sec:conclusions}

This paper presented a standardized benchmark for face forgery detection under realistic image degradation using the clean and degraded partitions of \gls*{mffi}.
Across six architecture families and up to seven input representations, clean-set performance proved to be an unreliable proxy for performance under degradation.
Xception achieved the strongest clean result (\major{\gls*{auc} $=0.884$}), but its performance decreased substantially after compression, resizing, and blurring.
Frozen DINOv3 exhibited the opposite profile, obtaining the highest degraded-set result~($0.726$) while requiring only the training of a linear classification~head.

The representation analysis showed that Fourier-domain information is most useful as an auxiliary cue.
Hybrid spatial-spectral inputs can improve convolutional models under degradation, whereas purely spectral representations consistently underperform \gls*{rgb}.
The qualitative attribution analysis further suggests that convolutional detectors emphasize localized facial evidence, while DINOv3 distributes evidence across broader facial structure.
Together, these findings demonstrate the importance of standardized degraded evaluation protocols and indicate that self-supervised visual representations are a promising basis for robust face forgery detection\major{, subject to the limitations discussed in \cref{sec:limitations}}.

\major{Future work should extend the evaluation to FaceForensics++~\cite{ref_faceforensics}, Celeb-DF~\cite{li2020celeb}, and DFDC~\cite{dolhansky2020deepfake} to distinguish robustness to image degradation from generalization across forgery sources.
Comparing frozen and fine-tuned pretrained backbones would clarify whether the observed robustness advantage persists after task-specific adaptation and provide a fairer comparison across model families.
Reporting results separately for each degradation type would identify the transformations to which each representation is most sensitive.
Finally, degradation-aware training and evaluation on diffusion-based forgeries would test whether the observed patterns hold for newer generative pipelines. The supervision-scale gap in \cref{tab:literature} also suggests that broader training distributions and ensembles of pretrained backbones may help narrow the remaining performance gap.}

\section*{\uppercase{Acknowledgments}}

\iffinal
    The authors thank the \textit{Pontifícia Universidade Católica do Paraná}~(PUCPR) for the financial support that made their participation in the conference possible.
\else
    \noindent\textit{The acknowledgments are hidden for review. The space below is reserved for the acknowledgments in the final version.}
    \vspace{2\baselineskip}
\fi

\bibliographystyle{IEEEtran}
\bibliography{bibtex}

\end{document}